# A multi parallel mixed-model disassembly line and its balancing optimization for fuel vehicles and pure electric vehicles

Qi Wang, Qingtao Liu*, Jingxiang Lv*, Xinji Wei, Jiongqi Guo,

Panyu Yu, Yibo Guo

Key Laboratory of Road Construction Technology and Equipment of MoE, Chang'an University, Xi'an, 710064, China

* Corresponding author.
E-mail address: qtaoliu@chd.edu.cn (Qingtao Liu); lvjx@chd.edu.cn (Jinxiang Lv)

**Abstract**: With the continuous growth of the number of end-of-life vehicles and the rapid increase in the ownership of pure electric vehicles, the automobile disassembly industry is facing the challenge of transitioning from the traditional fuel vehicles to the mixed disassembly of fuel vehicles and pure electric vehicles. In order to cope with the uncertainty of recycling quantity and the demand of mixed-model disassembly of multiple vehicle types, this paper designs a multi-parallel mixed-model disassembly line (MPMDL), and constructs a corresponding mixed-integer planning model for the equilibrium optimization problem of this disassembly line with the optimization objectives of the minimum number of workstations, the minimum fatigue level of workers and the minimum energy consumption. Combining the differences in disassembly processes between fuel vehicles and pure electric vehicles, an improved non-dominated sorting multi-objective genetic algorithm (INSGA-III) based on the distribution of feasible solutions and dynamic search resource allocation is designed to solve this multi-objective dynamic balance optimization problem, and the two-stage dynamic adjustment strategy is adopted to realize the adaptive adjustment of the disassembly line under the uncertainty of the recycling quantity, and, recently, arithmetic validation is carried out. The results show that the proposed method can effectively improve the resource utilization efficiency, reduce energy consumption, alleviate the workers' load, and provide multiple high-quality disassembly solutions under the multi-objective trade-off. Compared with mainstream multi-objective optimization algorithms, the INSGA-III algorithm shows significant advantages in terms of solution quality, convergence and stability. This study provides a green, efficient and flexible solution for hybrid disassembly of fuel and pure electric vehicles.

**1. Introduction**

　　Automobile recycling and disassembly are of strategic importance in promoting the development of a circular economy and environmental protection. Through the recycling of materials and parts from end-of-life vehicles, not only can it effectively reduce the demand for raw material extraction and processing and lower energy consumption, but it also provides strong support for the realization of the strategic goal of "carbon peak, carbon neutral" and promotes the transformation of the economy in the direction of green and low-carbon. In recent years, the number of end-of-life automobiles has continued to grow globally. Taking China as an example, data show that the overall number of end-of-life vehicles recycled in the country has been on a steady upward trend between 2014 and 2023, especially in 2022 and 2023, with significant year-on-year growth of about 28.9% and 25%, respectively, as shown in Fig.1(a). Meanwhile, the rapid development of pure electric vehicles has also brought about profound changes in the structure of the automobile industry. The national pure electric vehicle ownership increased from 80,000 units in 2014 to 14.01 million units in 2023, and the new ownership is nearly doubled in 2023, as shown in Fig.1(b). With the continuous climb in ownership, pure electric vehicles will occupy an increasingly important position in the future global automotive market, and their recycling and disassembly has become a key link in resource recycling and environmental governance, and the industry as a whole is facing the challenge of transitioning from the disassembly mode dominated by fuel vehicles to the parallel disassembly of fuel vehicles and pure electric vehicles(Zhou, 2023). In order to cope with the increasingly complex structure of end-of-life vehicles in the future, enterprises urgently need to build a flexible disassembly system compatible with the two types of vehicles, so as to achieve the goal of efficient, safe and green resource reuse. However, at present, the auto disassembly industry is still dominated by the disassembly of traditional fuel vehicles, and the relevant technical system is still difficult to meet the specialized and standardized demand for the disassembly of pure electric vehicles. Most automobile disassembly enterprises do not yet have the ability to deal with end-of-life pure electric vehicles. For example, as of June 2025, the number of qualified end-of-life motor vehicle recycling and disassembly enterprises in China is 1,800 (including traditional fuel vehicle and pure electric vehicle disassembly enterprises), and less than 1.5% of the enterprises have the qualification to dismantle pure electric vehicles.

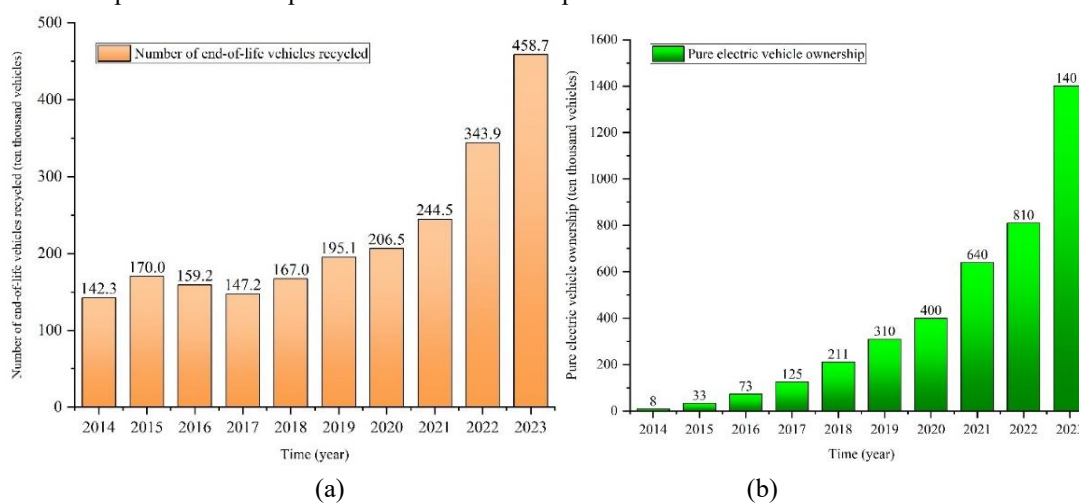

Fig. 1 (a)Number of end-of-life vehicles recycled in China in the past decade; (b) Ownership of purely electric vehicles in China in the past decade

The disassembly of a pure electric vehicle differs from that of a fuel vehicle in many ways. The pretreatment process for traditional fuel vehicles mainly includes disassembly the low-voltage battery, fuel tank, oil filter, steering wheel, three-way catalytic converter, wheels, as well as extracting fluids and detonating airbags. After disassembly the low-voltage battery of the pure electric vehicle, it is still necessary to remove the maintenance switch and wait for about 15 minutes to ensure the safety of the high-voltage system, and the disassembly of the battery pack can only be carried out after completing the discharging of the power battery to ensure that there is no danger before proceeding to the disassembly stage of the whole vehicle.

From the overall process, the traditional fuel car pretreatment is completed in order to carry out exterior and interior disassembly, and finally dismantle the engine, transmission, generator, steering system, braking system and suspension system. While pure electric vehicles are not very different from fuel vehicles in terms of exterior structure, their internal structure is very different. The electric drive system mostly adopts rear-mounted rear-drive design, while the front nacelle integrates key components such as high-voltage shunt box, DC/DC converter, low-voltage battery, on-board charger and vehicle controller. In addition, the time, tools, and safety standards required for disassembly pure Electric vehicles are significantly different from those for fuel vehicles(Wang et al., 2023). In this regard, the Ministry of Commerce of China has clearly pointed out in the Technical Requirements for the Recycling and Disassembly of End-of-Life Electric vehicles that enterprises intending to carry out the recycling and disassembly of end-of-life electric vehicles are required to carry out the corresponding technological transformation and equipment upgrading to enhance the recycling and disassembly capacity of pure electric vehicles.

Currently, pure electric vehicles are still in the early stage of end-of-life, and their recycling quantity is relatively small and highly volatile; in contrast, the recycling system of traditional fuel vehicles is more mature, but their recycling quantity is also affected by seasonal changes in the market. In contrast, the recycling system of traditional fuel vehicles is more mature, but its recycling quantity is also affected by seasonal changes in the market. When the recycling quantity is too high, enterprises need to bear high inventory holding costs; when the recycling quantity is insufficient, they will face inventory shortage costs. The uncertainty of the recycling quantity will directly affect the decision-making of enterprises in the design and construction of disassembly lines, which in turn affects their economic efficiency.

In actual operation, if an enterprise sets up a separate pure electric vehicles disassembly line, when the number of recycled vehicles is at an initial stage or when the growth is still unstable, it is very likely to cause problems such as low capacity utilization and increased equipment idleness, which will in turn affect the overall operational efficiency and economic benefits of the enterprise. On the contrary, as the scale of pure electric vehicle recycling continues to expand, the original disassembly line may face capacity bottlenecks, making it difficult to meet the requirements of operational tempo and disassembly efficiency. Therefore, when constructing and planning disassembly production lines, enterprises need to take into account the current and future changes in recycling demand, and have a certain degree of flexible expansion capability and the ability to adapt to vehicle models.

To this end, a highly flexible and schedulable multi-parallel hybrid disassembly line is proposed, which can realize a smooth transition from the disassembly of traditional fuel vehicles to pure electric vehicles. This hybrid disassembly line can operate stably under different models and different recycling quantities, effectively avoiding economic losses and resource wastage due to overcapacity or resource redundancy. At the same time, this disassembly line should also support the rapid realization of dynamic adjustment of the disassembly line configuration at different stages in order to enhance the responsiveness of the enterprise to the changes in the recycling market and ensure the continuity of the disassembly tasks.

The main contributions of this study are as follows:

(1) The idea of mixed-model disassembly of fuel vehicles and pure electric vehicles is proposed, and a multi-parallel mixed-model disassembly line (MPMDL) layout is designed, based on which a mixed-integer planning model under cooperative disassembly line layout is established.

(2) A dynamic adjustment strategy is proposed to solve the adaptive adjustment of disassembly lines under uncertain recycling quantity and the dynamic balance optimization of fuel vehicles and pure electric vehicles.

(3) A multi-objective evolutionary algorithm (INSGA-III) with equivalent task sets considering feasible solution distribution and real-time search resource allocation is designed.

## 2. Literature research

One of the core challenges facing the hybrid disassembly of fuel vehicles and pure electric vehicles is the disassembly line balance optimization problem (DLBP), which was first proposed by Prof. Gupta in 1999, and its research systematically reveals the essential differences between the disassembly line and the traditional assembly line in terms of structure and operation, and the core of the DLBP lies in the following: under the premise of following the constraints and priorities between the parts and components, disassembly tasks are reasonably distributed to the workstations to achieve a balanced distribution of workloads, thus improving the overall efficiency and resource utilization efficiency of the disassembly line. The core of DLBP is to reasonably allocate the disassembly tasks to each workstation and realize the balanced distribution of workload, so as to improve the overall efficiency of the disassembly line and the efficiency of resource utilization. Since the disassembly process is much higher than the assembly process in terms of task complexity, uncertainty and product heterogeneity, the proposal of DLBP has quickly attracted extensive attention from the academic community, and become an important research hotspot in the field of green manufacturing and sustainable recycling.

The primary problem affecting DLBP is the large amount of uncertainty in the disassembly system. In the actual disassembly process, the uncertainty of the damage degree of the parts often leads to the failure of the disassembly task, which becomes an important factor affecting the disassembly efficiency and the balance of the disassembly line. (Gungor and Gupta, 2001) discussed the various scenarios in which the task failure occurs, and defined the disassembly line balancing problem in this context as the disassembly line balancing problem with task failure, and proposed a corresponding solution based on network theory. To enhance the ability of the disassembly line to adapt to task failures, (Altekin and Akkan, 2012) proposed a predictive response method based on mixed integer programming, where a predictive balancing scheme is constructed in the first stage, and if the task fails, the failed task is readjusted and reassigned to the workstation in the second stage. Further, for certain parts with low recycling value and frequent task failures, (Altekin et al., 2008) introduced the concept of partial disassembly line equilibrium problem and proposed an optimization model based on linear programming relaxation with upper and lower bound constraints with the objective of profit maximization. (Bentaha et al., 2015) extended their study to scenarios containing partially disassembled and hazardous parts, where the task time is considered as a known normally distributed random variables, and is solved using with/or graphs modeling the task priority relationship, combined with second-order cone programming and convex segmented linear approximation techniques. In addition, the uncertainty of the disassembly order equally affects the disassembly efficiency and disassembly line balancing, (Kalayci and Gupta, 2013b) investigated the order-dependent disassembly line balancing problem and proposed a particle swarm algorithm based on the neighborhood variational operator. They then developed an artificial bee colony based algorithm (Kalayci and Gupta, 2013a), a taboo search based algorithm (Kalayci and Gupta, 2013c), and a hybrid algorithm combining a genetic algorithm with a variable neighborhood search (Kalayci et al., 2014) to improve the solution efficiency under different objectives and product complexities.

Even if the disassembly sequence is fixed, the uncertainty of the task time still exists. (Bentaha, Mohand Lounes et al., 2014b) for the first time considered the task time as a random variable with a known probability distribution, introduced a stochastic integer programming model, and combined Monte Carlo sampling techniques with the L-shape algorithm with the objective of minimizing the cost to solve the problem. In a follow-up study, they proposed a stochastic optimization model with the objective of profit maximization (Bentaha, M. Lounes et al., 2014) and a complex constraint model considering the presence of hazardous components (Bentaha, Mohand Lounes et al., 2014a), respectively, both of which used with/or graph modeling and segmented convex function approximation techniques. In terms of fuzzy uncertainty, (Kalayci et al., 2015) proposed a DLBP extension model for task processing time fuzzy, described the task time in terms of a triangular fuzzy affiliation function, and designed a hybrid discrete artificial bee colony algorithm to solve it. And in the scenario where the task cannot be completed in cycle time, (Altekin et al., 2016) proposed a hybrid remediation mechanism combining line stopping and offline disassembly, and developed a mixed integer programming model to maximize the expected profit.

For more complex uncertain environments, such as scenarios where partial disassembly and hazardous parts coexist with stochastic task times, (Bentaha et al., 2018) propose comprehensive profit-oriented disassembly line design and equilibrium models combining L-algorithms and Monte Carlo sampling for exact solutions, while (Xiao et al., 2020) transform the problem into a nonlinear robust integer programming model, which is further equivalently transformed into a linear model and designed an improved migratory bird optimization algorithm for efficient solution. In addition to the uncertainty of task and time, the uncertainty of the type, quality and quantity of parts also poses a challenge to the design of disassembly line. (Zeng et al., 2023) proposed an optimization method based on hybrid disassembly model, established a mixed integer programming model, focused on optimizing the objectives of the number of workstations, balancing, energy consumption and profit, and utilized the improved discrete flower pollination algorithm to solve the problem of insufficient number of batch disassembly due to the fine classification. The problem of insufficient disassembly quantity. Uncertainty exists not only in disassembly, but also in other parts of the remanufacturing system. (Liu et al., 2024) proposed a time evaluation method based on the fuzzy graph review technique for the uncertainty of assembly task time and combined it with adaptive two-layer genetic algorithm to optimize the assembly balance. (Li et al., 2020) In this study, based on the uncertainty theory to model the uncertainty of the assembly task time. Incompatible task set constraints are introduced and a simulated annealing algorithm with problem characteristics is designed to optimize the objective. All these uncertainty research methods provide a reference for solving the disassembly uncertainty problem. For the mixed disassembly problem of fuel and pure electric vehicles, the uncertainties existing in the traditional disassembly line also exist, and more critically, the uncertainty in the quantity of recycling. (Tang and Li, 2012) systematically reviewed the impact of uncertainty in the quantity and quality of returned goods in reverse logistics on the manufacturing system, including supply randomness, quality fluctuation, and path flexibility, which provides important insights for subsequent studies.(Gholizadeh et al., 2022) combined stochastic optimization with operator heuristic algorithms for multi-level, multi-product, and long-period systems to deal with recall quantity and cost uncertainty. (Niknejad and Petrovic, 2014) used fuzzy trapezoidal numbers to model recall quantity and quality uncertainty and optimized the network with a mixed integer algorithm.

In recent years, with the rapid growth of the number of end-of-life vehicles, the impact of automobile disassembly on energy consumption and environmental pollution has become more and more significant, so it is necessary to introduce environmental and social optimization objectives into the study of disassembly line balancing problem. (Güngör and Gupta, 2002) proposed a simple disassembly line balancing problem, which takes into account the idle time of the workstations, the demand for the product parts, the hazardous level of the parts, and the direction of disassembly. factors and solved using a heuristic algorithm; (Ren et al., 2018) proposed a three-stage solution method for a multi-objective DLBP that includes economic and environmental factors as well as disassembly line efficiency. This method can deal with the weight dependency between objectives in multi-criteria decision making and can effectively obtain a high-quality solution set. In order to achieve the goal of low carbon and environmental protection, (Yang et al., 2019) proposed a multi-objective optimization algorithm for disassembly sequence of used agricultural machines, the fruit fly optimization algorithm, which is oriented to low-carbon design for solving the multi-objective DLBP with the optimization goals of low carbon emission, low energy consumption, and low cost, and (Wang et al., 2019) constructed a multi-objective DLBP that takes into consideration the environmental impact and economic benefits of partial disassembly line evaluation index system, which requires all hazardous tasks to complete disassembly and optimizes the benefit indicators such as the number of workstations, workload smoothness, and disassembly profit, and proposes a new multi-objective genetic simulated annealing algorithm for solving it; (Budak, 2020) proposes a multi-periodic, multi-objective mixed-integer nonlinear programming model in terms of the three dimensions of economy, environment, and society and solved it using an improved augmented generalized Epsilon constraint method; (Yin et al., 2023), on the other hand, for the first time, proposed a partial disassembly line balancing problem by taking two key factors, namely, scrap end-of-life status and workers' skill differences, into consideration, constructed a mixed integer planning model, and improved NSGA-II by introducing incentive strategies to optimize the large-scale partial DLBP.

With the increasing number of disassembly, the traditional linear disassembly line is difficult to meet the actual demand. Therefore, scholars have successively introduced a variety of new disassembly line layouts such as parallel disassembly lines, bilateral disassembly lines, and U-shaped disassembly lines to enhance the disassembly efficiency. (Hezer and Kara, 2014) firstly investigated the single-product parallel disassembly line equilibrium problem and proposed a network solution method based on the shortest path model. However, this model fails to fully consider the complex execution constraints in the disassembly process and the additional energy consumption brought by dangerous tasks. To address the above shortcomings, (Liang et al., 2021) introduced the bilateral disassembly line balancing problem under complex execution constraints, established an improved mixed-integer programming model, and proposed a multi-objective dual-individual simulated annealing algorithm for solving the problem by jointly optimizing the objectives of weighted task length, workload balance, and total energy consumption. Further, (Liang et al., 2022) considered the adverse effects of possible destructive operations in the disassembly process, constructed a probabilistic task-destructive disassembly model based on probability, systematically analyzed its negative impacts on disassembly cost and workload smoothness, and proposed a multi-objective restarted genetic flatworm algorithm for the optimization of the two-side disassembly line balancing problem.

As the research continues to deepen, the optimization objectives and disassembly line types become more numerous, which contributes to the complexity of the model structure. In this regard, (Altekin, 2016) studied the stochastic disassembly line equilibrium problem with uncertain task time, assumed that the task time is a random variable obeying normal distribution, and constructed a segmented linear mixed integer planning model based on chance constraints with the objective of minimizing the number of chemical digits, which achieved better solution results than the traditional model. In a follow-up study, (Altekin, 2017) further proposed two second-order cone planning models and five segmented linear mixed integer planning models, and systematically compared the computation time and solution efficiency of the seven models, aiming at screening the modeling method with optimal performance. (Edis et al., 2019), constructed a generalized mixed-integer linear programming model, and validated its generalization and validity.

The above models are high in solution quality, but their modeling and computational processes are relatively complex, resulting in a long time-consuming algorithmic solution process. Therefore, scholars have gradually shifted their research focus to the optimization of the solution algorithms. (McGovern and Gupta, 2005) formally defined the demolition line equilibrium problem and proposed a genetic algorithm for obtaining the optimal or near-optimal solution. Based on this, (McGovern and Gupta, 2006) proposed a adaptive algorithm based on ant colony optimization for solving the disassembly line balancing problem, and (Pistolesi et al., 2018) developed the extreme multi-objective genetic algorithm (EMOGA) for efficiently solving the disassembly line balancing problem with multi-objective features.

In terms of disassembly line intelligence and automation, the concept of human-machine collaborative disassembly has been proposed to achieve a more efficient and safe operation process. (Xu et al., 2021) addressed the safety issue in the human-machine collaborative process, studied the human-machine collaborative disassembly information model considering the safety strategy, and proposed an improved discrete bee algorithm for solving the equilibrium problem of human-machine collaborative disassembly production line, which effectively improves the system's coordination and stability of the system. In addition to the consideration of safety, the actual disassembly process also needs to consider the tool replacement problem during the operation of industrial robots. For this reason, (Zeng et al., 2022) constructed a multi-objective equilibrium sequence model for the robot disassembly line, and modeled the relevant constraints and objectives. The model takes minimizing disassembly completion time, optimal disassembly energy consumption and profit maximization as the optimization objectives, and proposes an improved genetic simulated annealing algorithm for solving the problem, which verifies its high efficiency and adaptability under complex disassembly tasks.

Facing the disassembly line equilibrium problem under the influence of multiple uncertainties, there is an urgent need to carry out research on the systematic dynamic equilibrium mechanism. Scholars should consider different disassembly line types, mixed production modes, partial product demand, and resource supply constraints to construct a more realistic optimization model (Deniz and Ozcelik, 2019). Therefore, based on the uncertainty of recycling quantity, this paper designs a multi-parallel mixed-model disassembly line (MPMDL), constructs the corresponding mixed-integer planning model, and proposes an improved non-dominated sorted multi-objective genetic algorithm (INSGA-III) to solve the equilibrium problem of this type of disassembly line.

## 3. Problem description and model construction
### 3.1 Problem description

Based on the trend of disassembly capacity and recycling volume of the disassembly line, a reasonable disassembly schedule is formulated, from which the target production beats of the mixed-model disassembly line for fuel vehicles and pure electric vehicles are derived. Then, a multi-parallel mixed-model disassembly line (MPMDL) is designed by combining the respective disassembly prioritization matrices of fuel vehicles and pure electric vehicles to construct a disassembly task allocation model centered on the production beat constraint. This disassembly line aims to coordinate beat matching and task synergy among three or more parallel disassembly lines during the hybrid disassembly task allocation process. Specifically, in this paper, the MPMDL is modeled as a mixed-model disassembly system containing three parallel disassembly lines and two columns of shared workstations in full disassembly mode. As shown in Fig. 2, the three disassembly lines are arranged in parallel, among which the center is the main mixed-model disassembly line, which can disassemble fuel vehicles and pure electric vehicles at the same time, and the two sides of the line are responsible for disassembling specific vehicle models respectively. Shared workstations are set up between the parallel lines to take over the work assigned by the left and right disassembly lines, realizing dynamic deployment and sharing of resources. In the figure, the green tasks represent the disassembly tasks of pure electric vehicles, and the yellow tasks correspond to the disassembly tasks of fuel vehicles, which fully reflects the core idea of multi-vehicle mixed-model disassembly and resource synergy.

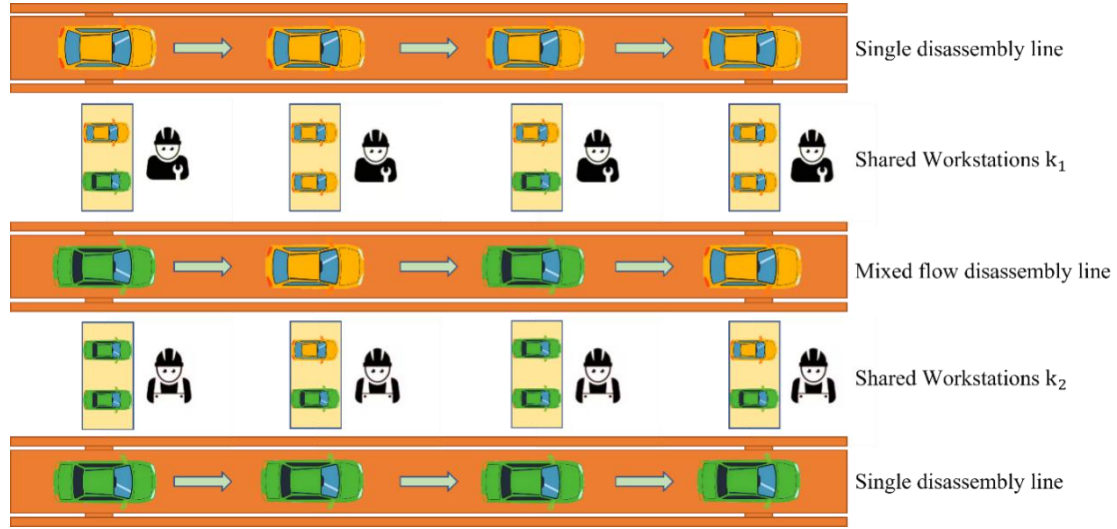

Fig. 2 Schematic layout of multi-parallel mixed-model disassembly line

### 3.2 Model assumptions and notation

To facilitate the mathematical description of the model, the following assumptions are given at this stage of the paper:

(1) The disassembly time of each task is known.
(2) The multi-parallel disassembly line consists of three parallel disassembly lines.
(3) The disassembly process is complete disassembly, i.e., all parts need to be disassembled.
(4) The resource constraints of the disassembly tools are not considered.
(5) Each worker can only work on one side of a disassembly line at a time.

In order to realize the dynamic balance optimization of fuel and pure electric vehicles under multiple parallel mixed-model disassembly lines, the symbols and variables related to the model are listed in Table 1.

**Table 1**

**Setting model variables and symbols**

| variable symbol | variable meaning | relations of satisfaction |
|---|---|---|
| $i$ | Index of types of retired automobiles | $0<i\leq N$ |
| $j$ | Index of months | $1\leq j\leq 12$ |
| $k$ | Index of remanufacturing, reuse and repurposing recycling businesses | $1\leq k\leq 3$ |
| $o$ | Index of disassembled lines | $1\leq o\leq 3$ |
| $N$ | Number of retired vehicle types | $N>0$ |
| $i_o, j_o$ | Index of disassembly tasks for disassembly line $o$ | $1\leq i_o, j_o\leq N_o$ |

| $N_o$ | Number of disassembly tasks for disassembly line o | $0<N_o$ |
|---|---|---|
| $P_{i_o j_o}$ | Prioritized relationship matrix for disassembly of Article o disassembly line | $P_{i_o j_o}=[p_{i_o j_o}]_{N_o \times N_o}$ |
| $n$ | Row index of shared workstations | $1 \leq n \leq 2$ |
| $d_n$ | Column index of shared workstations in row $n$ | $1 \leq d_n \leq k_n$ |
| $k_n$ | Number of shared workstations turned on in row $n$ | $1 \leq k_n$ |
| $ST_{k_n}$ | Sum of disassembly times for shared workstations in row n, column $k_n$ | $0 \leq ST_{d_n} \leq CT$ |
| $CT$ | Cycle time of production beats | $CT>0$ |
| $t_{i_o}$ | Time of the $i$th disassembly task in disassembly line $o$ | $0<t_{i_o}$ |
| $l_o$ | Number of assigned disassembly tasks in disassembly line o | $\sigma_i>0$ |
| $C_{k_n}$ | Set of tasks for shared workstation $k_n$ | $C_{k_n} \neq \emptyset$ |
| $PS_{l_o}$ | Task number corresponding to loth column of feasible disassembly sequence | $C>0$ |
| $ST_{K_n}$ | Sum of disassembly time and $t_{i_n}$ or $t_{i_{n+1}}$ for nth row and $k_n$th column of shared workstation | $C_s>0$ |
| $e_{i_o}$ | Energy consumption rate for the $i$th disassembly task of the $o$th disassembly line | $e_{i_o}>0$ |

The decision variables are as follows:

$$x_{i_0 d_n} = \begin{cases} 1, & \text{If the ith disassembly task of the oth disassembly line is assigned} \\ & \text{to the shared workstation of the nth row and } d_n \text{th column} \\ 0, & \text{if not} \end{cases}$$

### 3.3 Objective function

(1) Minimize the number of workstations

Reducing the number of workstations reduces some of the costs of the enterprise (e.g., workers' wages, equipment acquisition and maintenance costs, etc.) when the production beat of the disassembly line is kept constant. Therefore, one of the key objectives of optimizing the design of a disassembly line is to minimize the number of workstations while ensuring that disassembly efficiency and quality are not compromised.

$$f_1 = min\left(\sum_{n=1}^{2} k_n\right) \tag{4.1}$$

(2) Minimize staff fatigue

In the process of disassembly line design, if the optimization goal is only to minimize the number of workstations, it may lead to uneven distribution of the fatigue level of the operators at each workstation, thus affecting the quality and efficiency of the disassembly operation. To solve this problem, an energy consumption balance index can be introduced to quantitatively assess the fatigue level of the operators, thus realizing the dual optimization goals of workstation configuration and workload balance.

$$f_2 = min\left(\sum_{n=1}^{2}\sum_{d_n=1}^{k_n}\sum_{\substack{i_n=1,\\i_{n+1}=1}}^{N,N_{n+1}}\left(x_{i_n d_n} \cdot t_{i_n} \cdot e_{i_n} + x_{i_{n+1} d_n} \cdot t_{i_{n+1}} \cdot e_{i_{n+1}}\right)\right) / \sum_{n=1}^{2} k_n \tag{4.2}$$

(3) Minimize disassembly energy consumption

The energy consumption of the disassembly line mainly includes ventilation energy consumption per unit time $E_1$, lighting energy consumption $E_2$, task disassembly energy consumption $E_3$ and workstation standby energy consumption $E_4$. To address these energy consumption issues, the following improvements can be made: introducing energy-saving equipment, optimizing dismantling processes, and reducing standby energy consumption. This not only reduces operating costs, but also improves energy utilization efficiency and promotes the development of disassembly lines in a green and sustainable direction.

$$f_3 = \min \left( \begin{array}{c} \sum_{n=1}^{2} k_n \cdot CT \cdot (E_1 + E_2) + \sum_{n=1}^{2} \sum_{d_n=1}^{k_n} ST_{d_n} \cdot E_3 + \\ \left( \sum_{n=1}^{2} k_n \cdot CT - \sum_{\substack{i_n=1,\\ i_{n+1}=1}}^{N,N_{n+1}} \left( x_{i_n d_n} \cdot t_{i_n} + x_{i_{n+1} d_n} \cdot t_{i_{n+1}} \right) \right) \cdot E_4 \end{array} \right) \quad (4.3)$$

**3.4 Constraints**

The constraints for $f_1$, $f_2$ and $f_3$ are as follows:

$$\sum_{d_n=1}^{k_n} x_{i_n d_n} = 1, i_n = 1, 2, \cdots, N_n \quad (4.4)$$

$$ST_{d_n} \leq CT, d_n = 1, 2, \cdots, k_n \quad (4.5)$$

$$\sum_{i_o=1}^{N_o} x_{i_o d_o} \geq 1, d_o = 1, 2, \cdots, k_o \quad (4.6)$$

$$x_{i_o d_o} \in \{0, 1\}, \forall i_o, d_o \quad (4.7)$$

$$V \cup Z = N \quad (4.8)$$

$$\sum_{i_1=1}^{N_1} \sum_{d_1}^{k_1} x_{i_1 d_1} + \sum_{i_o=1}^{N_o} \sum_{d_o=1}^{k_o} x_{i_o d_o} + \sum_{i_o=1}^{N_o} \sum_{d_{o-1}=1}^{k_{o-1}} x_{i_o d_{o-1}} + \sum_{i_2=1}^{N_2} \sum_{d_2=1}^{k_2} x_{i_2 d_2} = N_o, o = 1, 2, 3 \quad (4.9)$$

Where Equation (4.4) indicates that each task must be and can only be assigned to one station; Equation (4.5) indicates that the total disassembly time of each shared workstation cannot be greater than the beat time; Equation (4.6) indicates that the open workstation cannot be empty, i.e., there exists at least one disassembly task at each workstation; Equation (4.7) indicates that the task assignment variables are decision variables, which take either 0 or 1; Equation (4.8) indicates that the sum of the assigned end-of-life car types and the end-of-life car types to be assigned is the total number of end-of-life car types; and Equation (4.9) indicates that the tasks on each disassembly line are assigned to shared workstations.

**4. Improved multi-objective genetic algorithm**

**4.1 Description of INSGA-III algorithm**

In a multi-parallel mixed-model disassembly line equilibrium optimization problem, all feasible solutions form a solution space tree with a branching structure. Due to the existence of multiple constraints, the number of feasible solutions in different subtrees varies significantly. Considering the distribution of feasible solutions, allocating more search resources to the subtree with more feasible solutions but fewer searches can effectively reduce the phenomenon of repeated searches and avoid a large amount of wasted search resources. In addition, due to the complex task prioritization constraints in the disassembly process of fuel vehicles and pure electric vehicles, the conventional crossover and mutation operations are very likely to generate infeasible solutions, which reduces the effectiveness of the algorithm. For this reason, this section proposes an improved multi-objective genetic algorithm (INSGA-III) with an equivalent task set that considers the distribution of feasible solutions and the real-time search resource allocation mechanism based on the NSGA-III algorithmic framework, which can enhance the solving capability and efficiency of the balance optimization problem of the complex mixed-model disassembly line.

**4.2 Coding**

This study first generates a disassembly prioritization relationship matrix based on the disassembly prioritization relationship matrix $P_{yx}=[p_{nm}]_{N_o \times N_o}$, where $N_o$ is the number of retired car tasks on the disassembly line $o$ and $p_{nm}$ is a variable consisting of [0,1]. If task n is the immediate predecessor of task $m$, then $p_{nm}=1$; otherwise, $p_{nm}=0$. The disassembly precedence relationship diagram is shown in Fig.3. Additionally, $P_{wg}$ represents the matrix of dangerous tasks and high-value demand tasks. In this matrix, the first row represents dangerous tasks, and when the value is 1, it indicates that the corresponding task is dangerous; the second row represents high-value demand tasks, and a value of 1 indicates that the task belongs to a high-value demand task. For example, the 1 in the fifth column of the first row indicates that the fifth task is a dangerous task, and the 1 in the third column of the second row indicates that the third task is a high-value demand task. Fig.4 shows the representation of the disassembly priority relationship matrix and the $P_{wg}$ matrix.

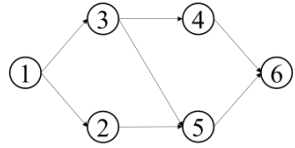

Fig. 3 Demolition prioritization relationship diagram

To describe the priority relationship between tasks, a priority matrix $P_{yx}=[p_{nm}]_{N_o \times N_o}$ is used to represent the priority order between task $n$ and task $m$.

$$p_{nm} = \begin{cases} 1, & \text{if task } n \text{ is the immediate predecessor of task } m; \\ 0, & \text{otherwise.} \end{cases}$$

$$P_{yx} = \begin{array}{c} \\ 1 \\ 2 \\ 3 \\ 4 \\ 5 \\ 6 \end{array} \begin{array}{c} 1 \quad 2 \quad 3 \quad 4 \quad 5 \quad 6 \\ \begin{bmatrix} 0 & 1 & 1 & 0 & 0 & 0 \\ 0 & 0 & 0 & 0 & 1 & 0 \\ 0 & 0 & 0 & 1 & 1 & 0 \\ 0 & 0 & 0 & 0 & 0 & 1 \\ 0 & 0 & 0 & 0 & 0 & 1 \\ 0 & 0 & 0 & 0 & 0 & 0 \end{bmatrix} \end{array}$$

(a) demolition prioritization relationship matrix

$$P_{wg} = \begin{bmatrix} 0 & 0 & 0 & 0 & 1 & 0 \\ 0 & 0 & 1 & 0 & 0 & 0 \end{bmatrix}$$

(b) hazardous and high-value demand task matrix

Fig.4 Demolition prioritization relationship matrix and hazardous and high-value demand task matrix

The coding process aims to construct a feasible sequence of disassembly tasks based on the disassembly line. In this paper, a task selection method based on a prioritization matrix is used to ensure that the generated individuals are all feasible solutions. The specific steps are as follows:

(1) According to the number of fuel vehicles and pure electric vehicles to be recycled, the three disassembly lines select the corresponding objects to be disassembled from fuel vehicles, pure electric vehicles and hybrid models, respectively, with the initialization step $i_0=1$;

(2) According to the priority relationship graph of the vehicle type selected by each disassembly line, the corresponding task priority relationship matrix $P_{yx}$ is generated;

(3) Randomly select a disassembly line $o$, and from its prioritization matrix, select all the tasks whose columns do not contain "1", i.e., all the tasks that have no prior dependency, to form a candidate task set $C$. The above figure is an example, and the candidate task set $C=1$;

(4) If the candidate task set $C$ contains only one task $x$, add it to the disassembly sequence set $R_o$, and skip to step (6); if the candidate set contains more than one task, execute step (5);

(5) Select a task x from within the candidate set $C$ to be added to the disassembly sequence set $R_o$ according to the following two heuristic rules. wherein, Rule 1: Prioritize the selection of hazardous tasks to reduce the safety risk during disassembly; and Rule 2: Prioritize the selection of tasks that have a remanufacturing value to increase the disassembly efficiency. If there exists more than one task that satisfies the above rules at the same time, one of them is randomly selected;

(6) Update the step to $i_0=i_0+1$. If all the tasks on the three disassembly lines have not yet been fully encoded, (i.e., there are still tasks that have not yet been added to the set of disassembly sequences $R_o$). At this time, change the elements of the priority matrix corresponding to the selected tasks whose rows and columns are 1 to 0, and return to step (3); otherwise, go to step (7);

(7) Output the final constructed full set of disassembly sequences $R_o$. For example, after the above priority matrix finishes encoding, the generated sequence of disassembly tasks for a certain feasible individual is: [1 2 3 4 5 6].

**4.3 Decoding**

The MPMDL balancing problem involves the assignment of tasks to the three decoding lines, where the shared workstations on both sides can execute the tasks assigned to the middle decoding line. In the decoding process, it is necessary to reasonably allocate tasks to the shared workstations and determine the execution order of tasks within each workstation. Since there are order constraints between tasks, it is necessary to ensure that the task assignments satisfy the preset sequential relationships. Based on this, the calculation expression for the workstation disassembly time $ST_{d_n}$ is shown below:

$$ST_{d_n} = \sum_{\substack{i_n=1, \\ i_{n+1}=1}}^{N_n, N_{n+1}} \left( x_{i_n d_n} \cdot t_{i_n} + x_{i_{n+1} d_n} \cdot t_{i_{n+1}} \right) \tag{4.10}$$

The main steps of decoding are as follows:

(1) Before decoding begins, all disassembly lines must be initialized. Set the current task to its initial state, i.e., $l_o=1$, $k_n=1$; set the initial disassembly time $ST_{d_n}=0$ for all shared workstations; in addition, the task set for all shared workstations is empty, i.e., $C_{k_n}=\emptyset$.

(2) Randomly select a decomposition line $o$. When $o=1$, search for $l_1$, $k_1$, $ST_{k_1}$ and $C_{k_1}$; when $o=2$ search for $l_2$, $k_1$, $k_2$, $ST_{k_1}$, $ST_{k_2}$, $C_{k_1}$, and $C_{k_2}$; when $o=3$ search for $l_3$, $k_2$, $ST_{k_2}$ and $C_{k_2}$.

(3) Let $i_o=PS_{l_o}$, When $o=1$, calculate the total disassembly time of task $i_1$ on shared workstation $k_1$, i.e., $ST_{K_1}=ST_{k_1} + t_{i_1}$. If $ST_{K_1} \leq CT$, assign task $i_1$ to shared workstation $k_1$ and update the total disassembly time and task set for that workstation, i.e., $ST_{k_1}=ST_{K_1}$, $C_{k_1}=C_{k_1} \cup \{i_1\}$; if $ST_{K_1} > CT$, proceed to step (5); When $o=2$, calculate the total disassembly time when task $i_2$ is assigned to shared workstations $k_1$ and $k_2$ respectively, denoted as: $ST_{K_1} = ST_{k_1} + t_{i_1}$, $ST_{K_2}=ST_{k_2} + t_{i_2}$. The following judgments are made: If $ST_{K_1} \leq CT$ and $ST_{K_2} > CT$, assign task $i_2$ to shared workstation $k_1$ and update $ST_{k_1} = ST_{K_1}$, $C_{k_1} = C_{k_1} \cup \{i_2\}$; If $ST_{K_1} > CT$ and $ST_{K_2} \leq CT$, assign task $i_2$ to shared workstation $k_2$ and update $ST_{k_2} = ST_{K_2}$, $C_{k_2} = C_{k_2} \cup \{i_2\}$; If $ST_{K_1} > CT$ and $ST_{K_2} \leq CT$, proceed to (4); If $ST_{K_1} > CT$ and $ST_{K_2} > CT$, proceed to step (5); When $o=3$, calculate the total disassembly time of task $i_3$ at shared workstation $k_2$, i.e., $ST_{K_2} = ST_{k_2} + t_{i_3}$. If $ST_{K_2} \leq CT$, assign task $i_3$ to shared workstation $k_2$ and update the total disassembly time and task set for that workstation: $ST_{k_2} = ST_{K_2}$, $C_{k_2} = C_{k_2} \cup \{i_3\}$; if $ST_{K_2} > CT$, proceed to step (5).

(4) To reduce the number of shared workstations that are turned on and their idle time, tasks should be prioritized and assigned to shared workstations that minimize their remaining time. Let the current task be $i_2$. Compare the remaining available time $CT - ST_K$ of each workstation. The specific rules are as follows: If $CT - ST_{K_1} > CT - ST_{K_2}$, assign task $i_2$ to shared workstation $k_2$ and update; If $CT - ST_{K_1} < CT - ST_{K_2}$, assign task $i_2$ to shared workstation $k_1$ and update; If $CT - ST_{K_1} = CT - ST_{K_2}$, randomly select a workstation to assign task $i_2$ to and update its disassembly time and task set.

(5) Open a new shared workstation. When $o=1$, add a new shared workstation, i.e., $k_1=k_1+1$; when $o=2$, randomly select one of the shared workstations $k_1$ and $k_2$ and increment its number by one; when $o=3$, add a new shared workstation $k_2$, i.e., $k_2=k_2+1$, then return to step (3) and continue with task allocation.

(6) Determine whether task allocation is complete. If the current number of tasks on the disassembly line, $l_o \neq N_o$ (i.e., there are still unassigned tasks on this disassembly line), set $l_o = N_o$ and return to step (2) to continue task allocation; if $l_o = N_o$, it indicates that all tasks on this disassembly line have been allocated, remove it from the candidate set, and repeat steps (2) to (6); if all disassembly lines satisfy $l_o = N_o$, proceed to step (7);

(7) Output the final disassembly task sequences for all shared workstations, i.e., the ordered allocation results of task sets within each shared workstation $C_k$.

These are the specific steps of decoding and the flowchart is shown in Fig. 5 below.

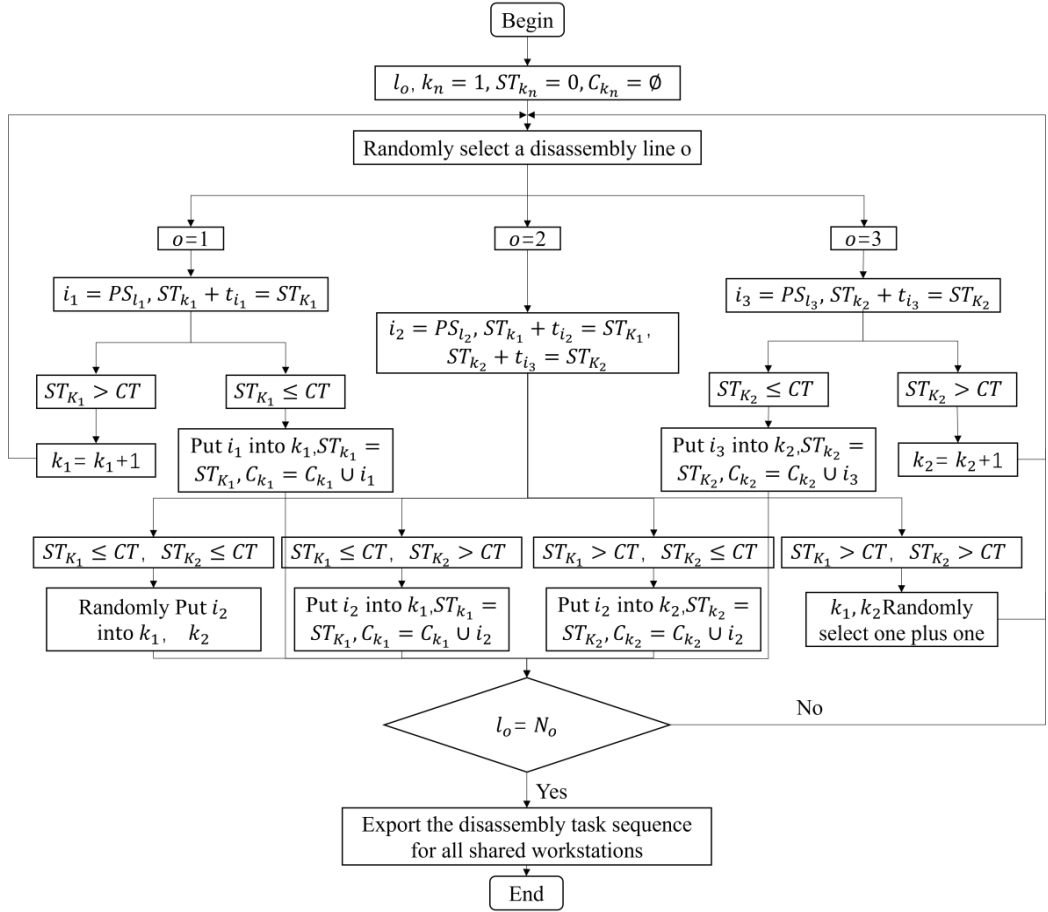

Fig. 5 Decoding flowchart

**4.4 Initialization, crossover and mutation**

During the initialization phase, this paper constructs three population sets: *popran*, *popzong*, and *popchun*, which store the initial disassembly information for gasoline vehicles, hybrid vehicles, and pure electric vehicles, respectively. To improve the quality of the initial disassembly, the concept of an equivalent task set is introduced. This set consists of disassembly tasks that satisfy the constraint conditions, can be regarded as equivalent, and can be scheduled arbitrarily. During each generation of the disassembly sequence, based on the current equivalent task set and historical initialization data, the number of times each node task has been accessed is counted, and in subsequent initializations, node tasks with lower historical access frequencies are prioritized to enhance the diversity and exploration capabilities of the solution.

To effectively retain high-quality individuals during the solution process, a single-point crossover strategy is adopted. However, due to the high structural dependency and complex priority constraints of mixed-model disassembly tasks, random crossover is highly likely to produce infeasible solutions. To address this issue, this paper designs an equivalent task set to store the set of tasks available for crossover.

First, individuals participating in crossover are selected based on crossover probability. Then, a non-empty cell array is randomly selected from the equivalent task set as the crossover reference. Next, the indices of the crossover tasks are extracted from *popran*, *popzong*, and *popchun*, respectively. When the index difference between two tasks is 1, the tasks are swapped to generate a feasible solution; if the index difference is greater than 1, all intermediate tasks between the two crossover tasks are extracted and appended to the task with the smaller index to satisfy the task sequence constraint, thereby ensuring the feasibility of the crossover result. This process is illustrated in Fig.6, ensuring that while maintaining algorithm convergence, the explorability of the solution space and the structural rationality of individuals are enhanced.

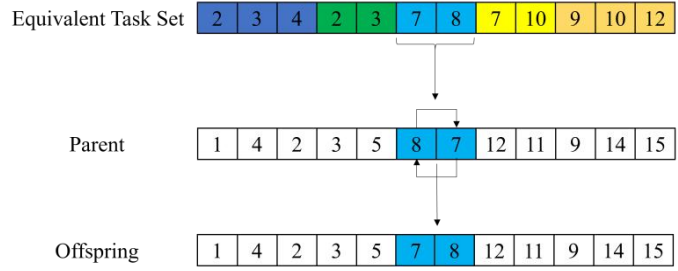

Fig. 6 Cross operator

Based on the predefined mutation probability pm, this paper employs the following strategy: first, individuals requiring mutation are selected from the current population according to the mutation probability. Subsequently, a non-empty cell array is randomly selected from the constructed equivalent task set as the candidate task set for mutation. Next, the index positions of the tasks to be mutated are extracted from *popran*, *popzong*, and *popchun*, respectively. To ensure that the mutated solution is feasible, the individuals following the smaller value of the mutated task index are initialized, thereby constructing a feasible mutated solution that satisfies the constraints, as shown in Fig.7.

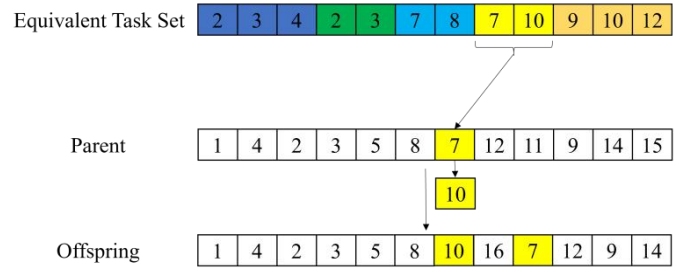

Fig. 7 Variational operators

**4.5 Dynamic Adjustment Strategies**

Given the increasing number of gasoline-powered vehicles and pure electric vehicles being recycled in disassembly companies, significant profit losses occur when recycling volumes fluctuate excessively. Therefore, this paper proposes a dynamic adjustment strategy, which is divided into two stages: the first stage applies when recycling volumes fluctuate significantly but do not affect the type of disassembly line used, and the second stage applies when recycling volumes fluctuate significantly and do affect the type of disassembly line used. The following adjustment schemes are proposed for each of these two stages.

First-stage adjustment scheme: When initializing future predicted recycling quantities, only the current recycling quantity within 25% is initialized. The method proposed in this paper is applied to calculate and optimize. When the prediction curve is slow, the time required for adjustment is longer; conversely, the time required for adjustment is shorter.

Second-stage adjustment plan: When the recycling quantities of gasoline vehicles and pure electric vehicles are very large, the disassembly types of the corresponding parallel mixed-model disassembly lines should also be adjusted. Some symbolic descriptions of this adjustment scheme are as follows. $DA_t$: the total optimal disassembly quantity of gasoline vehicles and pure electric vehicles; $DA_{fv}$: the optimal disassembly quantity of gasoline vehicles; $DA_{pev}$: the optimal disassembly quantity of pure electric vehicles; $DA_{sl}$: the monthly disassembly quantity of a single disassembly line; their

relationships are as follows:

$$DA_t = DA_{fv} + DA_{pev} = DA_{sl} * 3 \tag{22}$$

When $min(DA_{fv}, DA_{pev}) \leq DA_{sl}$:

If $DA_{fv} \leq DA_{sl}$, both sides of the disassembly line select pure electric vehicles for disassembly. If $DA_{pev} \leq DA_{sl}$, both sides of the disassembly line select fuel vehicles for disassembly.

When $DA_{sl} < min(DA_{fv}, DA_{pev}) < 2DA_{sl}$:

the two sides of the disassembly line one selects fuel vehicles for disassembly and one selects pure electric vehicles for disassembly.

## 5. Case studies

### 5.1 algorithm analysis

According to market research results, a certain company needs to dismantle an average of 1,354 vehicles per month (1,255 gasoline-powered vehicles and 99 pure electric vehicles). Considering the use of three parallel mixed-model disassembly lines operating simultaneously, the monthly disassembly task volume for a single disassembly line is approximately one-third of the company's total disassembly volume, or 451 vehicles. However, the monthly recycling volume of pure electric vehicles is currently far below the total disassembly capacity of a single line. Therefore, according to the dynamic adjustment strategy, pure electric vehicles should be dismantled on the middle mixed-model disassembly line, while both side lines dismantle gasoline-powered vehicles. The target production rhythm is calculated to be 650 seconds. Under this production rhythm constraint, tasks should be reasonably allocated among processes based on the disassembly priority relationship diagram for fuel-powered vehicles, pure electric vehicles, and mixed-model vehicles. The corresponding disassembly priority relationship diagram is shown in Fig.8.

The relevant parameters in the algorithm model are set as follows: maximum generation $maxgen = 20$, mutation probability $pm = 0.1$, crossing probability $pc = 0.8$, population size $n = 200$, number of objective functions $N = 3$, reference point score $M = 5$, ventilation energy consumption per unit time $E1 = 20$, lighting energy consumption per unit time $E2 = 40$ kWh, disassembly energy consumption per unit time $E3 = 90$ kWh, and standby energy consumption per unit time $E4 = 55$ kWh.

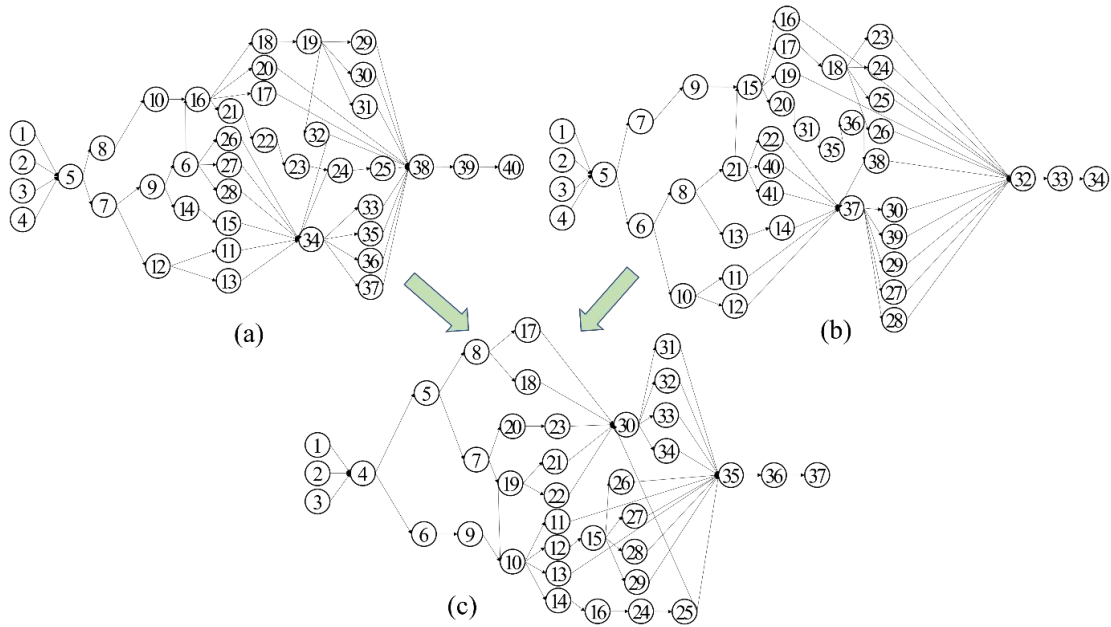

Fig. 8 (a) Fuel vehicle disassembly prioritization relationship diagram; (b) Pure electric vehicle disassembly prioritization relationship diagram; (c) Integrated disassembly prioritization relationship diagram

Table 2

**All Pareto solutions solved by INSGA-III algorithm**

| ID | $f_1$ | $f_2$ | $f_3$ |
|----|-------|-------|-------|
| 1 | 40 | 27 | 3241 |
| 2 | 39 | 25 | 3297 |
| 3 | 36 | 27 | 3460 |
| 4 | 37 | 26 | 3409 |
| 5 | 38 | 26 | 3341 |
| 6 | 41 | **24** | **3204** |
| 7 | **34** | 29 | 3641 |
| 8 | 35 | 28 | 3553 |

All pareto solutions solved by the algorithm are summarized in Table 2, with a total of 8 non-inferior solutions. Where $f_1$ represents the minimum number of workstations. Staff fatigue $f_2$ is quantified by worker energy consumption, and the solution is the average energy consumption of workers at a single workstation in *kcal*. $f_3$, the disassembly energy consumption, is quantified by electrical energy consumption, and the solution is the sum of electrical energy consumed by all workstations in *kwh*. Then, the Gantt charts are plotted for the time when $f_1$ is optimal, and the Gantt charts are plotted for the time when $f_2$ and $f_3$ are optimal, respectively.

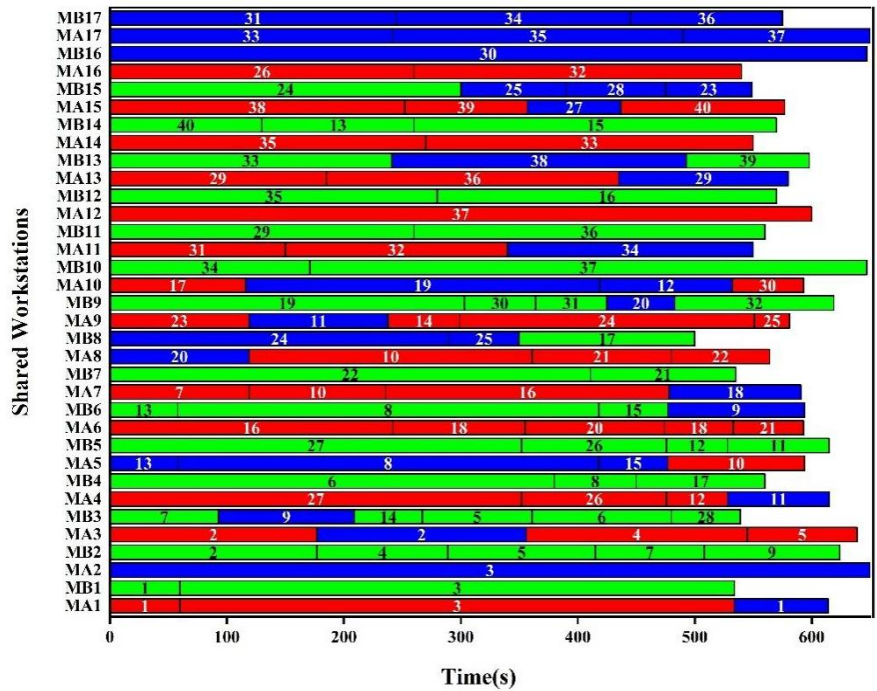

(a) Gantt chart at $f_1$ optimization

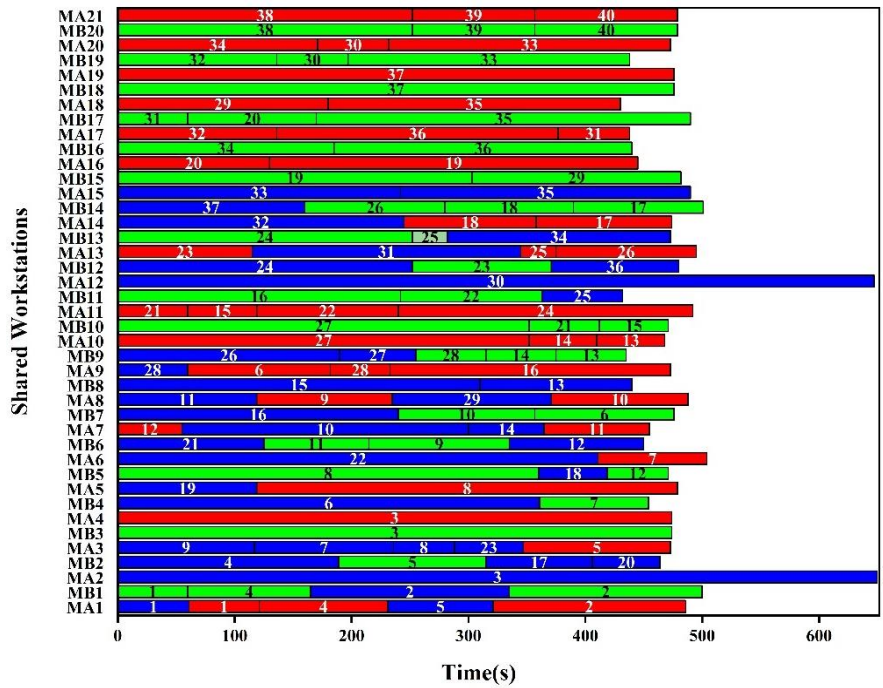

(b) Gantt chart at $f_2$, $f_3$ optimization

Fig. 9 Gantt chart corresponding to the optimal solution for the three target values

Fig.9(a) shows that under the optimal number of workstations configuration, the actual load time of each shared workstation is close to the target production cycle, idle time is effectively compressed, and time utilization is significantly improved. This is mainly because as the number of workstations decreases, the process allocation becomes more reasonable, thereby reducing the idle time of each workstation and improving overall operational efficiency. Therefore, the number of workstations at this time can be considered the optimal configuration.

In contrast, in Fig. 9(b), the load tasks are distributed across more shared workstations, resulting in increased idle time at each workstation. This configuration sacrifices some floor space utilization but provides operators with more rest time while improving ventilation conditions in the factory area, thereby reducing worker fatigue and ventilation-related energy consumption, and enhancing workers' sustained operational capacity. Therefore, enterprises can flexibly select the disassembly scheme that best aligns with their priority objectives from the Pareto front solution set based on their actual operational needs. The three-dimensional visualization distribution of all non-inferior solutions is shown in Fig.10.

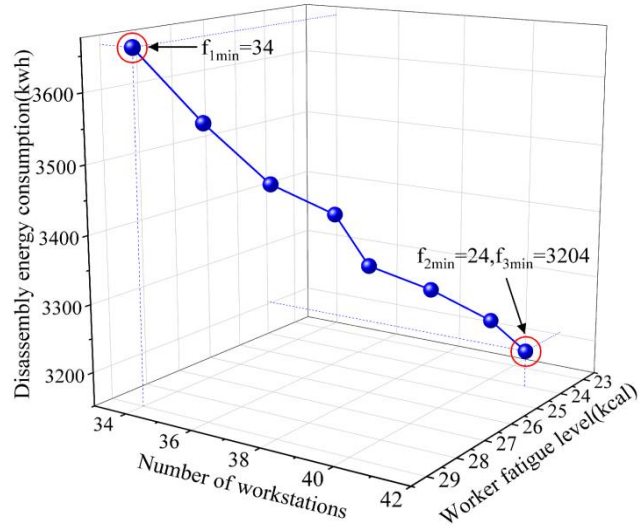

Fig. 10 Pareto front for the INSGA-III algorithm

Further verification can be obtained from the Gantt chart and Pareto distribution diagram: each disassembly scheme achieves an effective trade-off between multiple objectives, demonstrating the diversity and rationality of the solutions. This also indirectly proves the superiority and applicability of the improved non-dominated sorting genetic algorithm (INSGA-III) proposed in this paper in addressing complex mixed-model disassembly line balancing problems.

**5.2 Algorithm validation**

To validate the effectiveness and superiority of the INSGA-III algorithm proposed in this paper for solving multi-parallel mixed-model disassembly line balancing optimization problems, classic multi-objective optimization algorithms were selected for comparison and analysis, specifically including: multi-objective particle swarm optimization algorithm, multi-objective whale optimization algorithm, and non-dominated sorting genetic algorithm. To ensure the validity of the comparison, the maximum number of iterations for all algorithms was uniformly set to 100. Additionally, to comprehensively evaluate the adaptability and solution performance of the algorithms under different problem scales, the study cases were categorized into small, medium, and large scales, with specific classification criteria detailed in Table 3.

**Table 3**

**Multi-objective algorithm parameters.**

| Algorithm | Parameter | Parameter settings | | |
|---|---|---|---|---|
| | | small | medium | large |
| INSGA-III | Population size | 20 | 40 | 80 |
| | Crossover probability | 0.5 | 0.7 | 0.9 |
| | Mutation probability | 0.05 | 0.1 | 0.15 |
| MOPSO | Population size | 20 | 40 | 80 |
| | Individual learning factor | 0.8 | 1.5 | 3 |
| | Group learning factor | 0.8 | 1.5 | 3 |
| | Inertia factor | 0.5 | 0.8 | 1.2 |
| NSWOA | Population size | 20 | 40 | 80 |

| NSGA-III | Population size | 20 | 40 | 80 |
| --- | --- | --- | --- | --- |
| | Crossover probability | 0.5 | 0.7 | 0.9 |
| | Mutation probability | 0.05 | 0.1 | 0.15 |

Table 4
**Test results of four algorithms.**

| Algorithm | | | INSGA-III | MOPSO | NSWOA | NSGA-III |
| --- | --- | --- | --- | --- | --- | --- |
| small | $f_1$ | Max | 41 | 41 | 41 | 41 |
| | | Min | 35 | 36 | 35 | 35 |
| | | Ave | 38 | 38.5 | 38 | 38 |
| | $f_2$ | Max | 28 | 28 | 28 | 28 |
| | | Min | 24 | 24 | 24 | 24 |
| | | Ave | 26 | 26 | 26 | 26 |
| | $f_3$ | Max | 3542 | 3605 | 3611 | 3856 |
| | | Min | 3197 | 3218 | 3228 | 3241 |
| | | Ave | 3369.5 | 3411.5 | 3419.5 | 3548.5 |
| medium | $f_1$ | Max | 41 | 41 | 41 | 41 |
| | | Min | 34 | 35 | 35 | 34 |
| | | Ave | 37.5 | 38 | 38 | 37.5 |
| | $f_2$ | Max | 29 | 28 | 29 | 28 |
| | | Min | 24 | 24 | 24 | 24 |
| | | Ave | 26.5 | 26 | 26.5 | 26 |
| | $f_3$ | Max | 3605 | 3703 | 3670 | 3825 |
| | | Min | 3208 | 3238 | 3199 | 3223 |
| | | Ave | 3406.5 | 3470.5 | 3434.5 | 3524 |
| large | $f_1$ | Max | 41 | 41 | 41 | 41 |
| | | Min | 34 | 34 | 35 | 34 |
| | | Ave | 37.5 | 37.5 | 38 | 37.5 |
| | $f_2$ | Max | 29 | 28 | 28 | 28 |
| | | Min | 24 | 24 | 24 | 24 |
| | | Ave | 26.5 | 26 | 26 | 26 |
| | $f_3$ | Max | 3460 | 3823 | 4007 | 3622 |
| | | Min | 3193 | 3223 | 3233 | 3204 |
| | | Ave | 3326.5 | 3523 | 3620 | 3413 |

As can be seen from the table, the improved INSGA-III algorithm achieves better solutions than the other multi-objective algorithms under all the three scale parameters, and the disassembly energy consumption decreases significantly with the increase in the number of scales as compared to that of the pre-improved NSGA-III algorithm. In order to illustrate the efficiency of the INSGA-III algorithm more intuitively and adequately, we plot the Pareto frontiers of the four algorithms under the optimal scale parameters, and two performance metrics, namely, Hypervolume Metric (HV) and Inverse Generation Distance (IGD) are used to evaluate the convergence and diversity of the solution set, and the box-and-line diagrams of the HV and IGD are plotted as in Fig. 11.Since the test problem instance of the actual Pareto frontier is unknown, we unite all the solutions obtained by the algorithms and select all the non-dominated solutions as the true Pareto frontier.

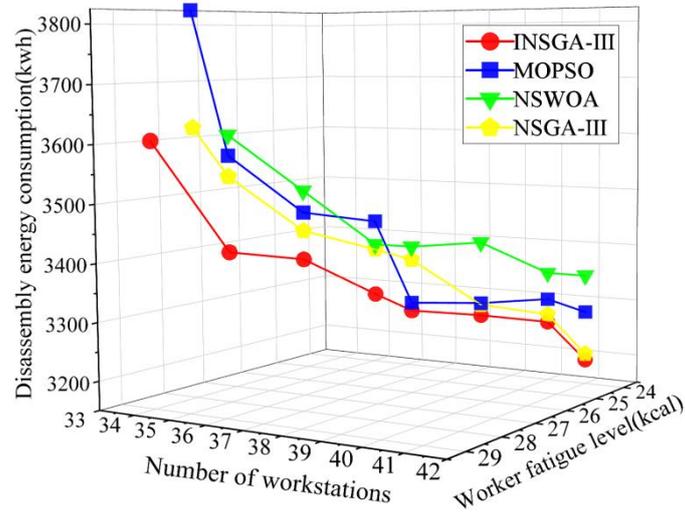

(a) Pareto frontier diagram

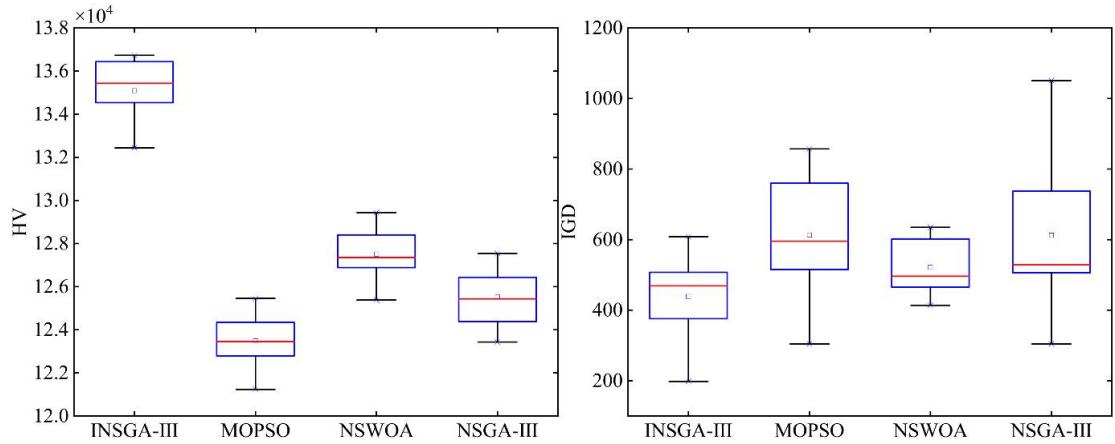

(b) box-and-line diagram

Fig. 11 Pareto and box-and-line diagrams for the four algorithms

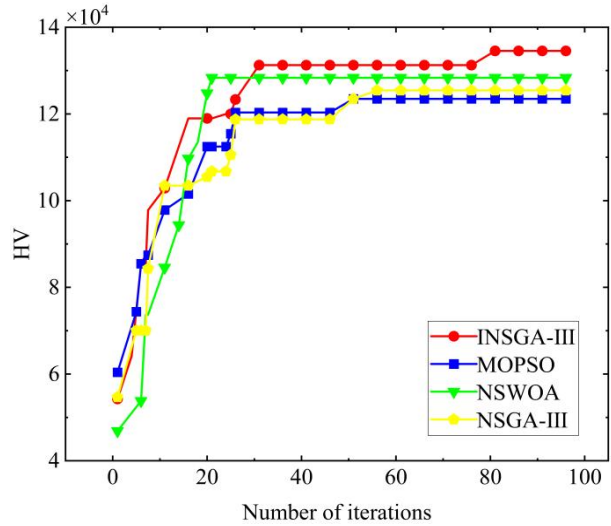

Fig. 12 Diagram of the iterative process of the four algorithms

As shown in Fig.11(a), the Pareto solutions obtained by INSGA-III exhibit a significantly superior overall distribution compared to the three comparison algorithms—MOPSO, NSWOA, and NSGA-III—particularly when the number of workstations is reduced, where its optimization advantage becomes even more pronounced. This phenomenon is primarily attributed to the high complexity of the solution space tree formed by gasoline vehicles and electric vehicles in the hybrid vehicle disassembly problem. During the initial solution generation phase, INSGA-III leverages an improved equivalent task set strategy and historical information guidance mechanism to rapidly and efficiently cover a wide range of feasible solution spaces. Additionally, during crossover and mutation processes, it enables precise search of feasible solution spaces, effectively avoiding the generation of infeasible solutions, thereby enhancing algorithm stability and optimization quality. In summary, INSGA-III demonstrates stronger capabilities in global solution space exploration, enabling it to achieve superior optimization results across multiple objective dimensions.

Fig.11(b) further validates the superiority of INSGA-III in multi-objective optimization performance. In the evaluation of the two key metrics, HV and IGD, the average values of INSGA-III are significantly better than those of the three comparison algorithms, and its fluctuation range is the smallest, indicating that the algorithm demonstrates higher stability and robustness in multiple independent runs. It can also be concluded that INSGA-III has the highest HV value and the lowest IGD value, confirming that the solution set obtained not only approaches the true Pareto frontier more closely but also has obvious advantages in terms of solution distribution breadth and convergence accuracy. To further analyze the convergence characteristics of the algorithm during the iteration process, Figure 12 shows the trend of changes in the HV metric for each algorithm. It can be seen that INSGA-III demonstrates rapid convergence capability in the early stages of iteration, quickly approaching the high-quality solution region. In contrast, while NSWOA exhibits strong local search capability in the middle stages of iteration, INSGA-III, with its superior ability to escape local optima in later iterations, successfully explores a better solution set and ultimately achieves the highest convergence level.

The comprehensive comparison results indicate that the INSGA-III algorithm not only exhibits faster convergence speed and broader solution coverage in the global search phase but also possesses the ability to escape local optima in the later stages of iteration, effectively avoiding the generation of infeasible solutions and significantly improving the feasibility of the solution and overall algorithm efficiency. In summary, the INSGA-III algorithm proposed in this paper demonstrates superior comprehensive performance when solving complex MPMDL balancing optimization problems, offering promising application prospects and promotional value.

## 6. Conclusions and future research

The MPMDL scheme proposed in this paper can effectively address the mixed dismantling of fuel-powered vehicles and pure electric vehicles under conditions of uncertain recycling quantities. Specifically, among the three parallel disassembly lines, the middle mixed-model line supports the parallel disassembly of fuel vehicles and pure electric vehicles, and the disassembly lines on both sides flexibly assign tasks according to the number of vehicle models. The two shared workstations set between the disassembly lines can effectively improve the utilization of operational resources and enhance the dynamic adaptability of the production line. The following main conclusions are obtained from the study:

(1) The designed dynamic adjustment mechanism can effectively cope with the fluctuation of the number of fuel vehicles and pure electric vehicles recycled, significantly reduce the idle rate of the production line resources, and improve the disassembly efficiency and flexibility.

(2) The INSGA-III proposed in this paper outperforms classic multi-objective optimization algorithms such as MOPSO, NSWOA, and NSGA-III in terms of solution quality, convergence speed, and stability.

(3) The obtained Pareto-optimal solution set exhibits good solution diversity, which can provide a variety of optional solutions for enterprises under different business objectives (energy saving, cost reduction or burden reduction).

As the end-of-life volume of pure electric vehicles continues to climb, hybrid disassembly of fuel-fired vehicles and pure electric vehicles may be the main mode of automobile disassembly in the future. Future research can further explore new production line layout forms (e.g., U-shaped disassembly line, etc.) to adapt to the mixed-model disassembly of multiple vehicle models, and try to integrate machine learning and heuristic algorithms to improve the intelligence level and solution efficiency of balance optimization of disassembly lines.

**CRediT authorship contribution statement**

**Qi Wang**: Writing-Computational experiments, Software, Data Curation. **Qingtao Liu**: Conceptualization, Methodology, Writing, Funding acquisition. **Jingxiang Lv**: Writing-Reviewing and Editing, Project administration. **Xinji Wei**: Writing- Modeling and Algorithm, Software, Validation. **Jiongqi Guo**: Enterprise investigation, Data collection. Panyu Yu: Conceptualization, Editing. **Yibo Guo**: Methodology, Editing.

**Declaration of competing interest**

The authors declare that they have no known competing financial interests or personal relationships that could have appeared to influence the work reported in this paper.

**Acknowledgments**

This research is partially supported by Xi'an Qin Chuangyuan's Innovation-Driven Platform Construction Special Project, grant number 21ZCZZHXJS-QCY6-0013 and the National Natural Science Foundation of China, grant number 52275005).

**References**

Altekin, F.T., 2016. A Piecewise Linear Model for Stochastic Disassembly Line Balancing. IFAC PapersOnLine 49(12), 932–937. https://doi.org/10.1016/j.ifacol.2016.07.895.

Altekin, F.T., 2017. A comparison of piecewise linear programming formulations for stochastic disassembly line balancing. International Journal of Production Research 55(24), 7412–7434. https://doi.org/10.1080/00207543.2017.1351639.

Altekin, F.T., Akkan, C., 2012. Task-failure-driven rebalancing of disassembly lines. International Journal of Production Research 50(18), 4955–4976. https://doi.org/10.1080/00207543.2011.616915.

Altekin, F.T., Bayındır, Z.P., Gümüşkaya, V., 2016. Remedial actions for disassembly lines with stochastic task times. Computers & Industrial Engineering 99, 78–96. https://doi.org/10.1016/j.cie.2016.06.027.

Altekin, F.T., Kandiller, L., Ozdemirel, N.E., 2008. Profit-oriented disassembly-line balancing. International Journal of Production Research 46(10), 2675–2693. https://doi.org/10.1080/00207540601137207.

Bentaha, M.L., Battaïa, O., Dolgui, A., 2014a. An exact solution approach for disassembly line balancing problem under uncertainty of the task processing times. International Journal of Production Research 53(6), 1807–1818. https://doi.org/10.1080/00207543.2014.961212.

Bentaha, M.L., Battaïa, O., Dolgui, A., 2014b. A sample average approximation method for disassembly line balancing problem under uncertainty. Computers & Operations Research 51, 111–122. https://doi.org/10.1016/j.cor.2014.05.006.

Bentaha, M.L., Battaïa, O., Dolgui, A., Hu, S.J., 2014. Dealing with uncertainty in disassembly line design. CIRP Annals 63(1), 21–24. https://doi.org/10.1016/j.cirp.2014.03.004.

Bentaha, M.L., Battaïa, O., Dolgui, A., Hu, S.J., 2015. Second order conic approximation for disassembly line design with joint probabilistic constraints. European Journal of Operational Research 247(3), 957–967. https://doi.org/10.1016/j.ejor.2015.06.019.

Bentaha, M.L., Dolgui, A., Battaïa, O., Riggs, R.J., Hu, J., 2018. Profit-oriented partial disassembly line design: dealing with hazardous parts and task processing times uncertainty. International Journal of Production Research 56(24), 7220–7242. https://doi.org/10.1080/00207543.2017.1418987.

Budak, A., 2020. Sustainable reverse logistics optimization with triple bottom line approach: An integration of disassembly line balancing. Journal of Cleaner Production 270. https://doi.org/10.1016/j.jclepro.2020.122475.

Deniz, N., Ozcelik, F., 2019. An extended review on disassembly line balancing with bibliometric & social network and future study realization analysis. Journal of Cleaner Production 225, 697–715. https://doi.org/10.1016/j.jclepro.2019.03.188.

Edis, E.B., Ilgin, M.A., Edis, R.S., 2019. Disassembly line balancing with sequencing decisions: A mixed integer linear programming model and extensions. Journal of Cleaner Production 238. https://doi.org/10.1016/j.jclepro.2019.117826.

Gholizadeh, H., Goh, M., Fazlollahtabar, H., Mamashli, Z., 2022. Modelling uncertainty in sustainable-green integrated reverse logistics network using metaheuristics optimization. Computers & Industrial Engineering 163. https://doi.org/10.1016/j.cie.2021.107828.

Gungor, A., Gupta, S.M., 2001. A solution approach to the disassembly line balancing problem in the presence of task failures. International Journal of Production Research 39(7), 1427–1467. https://doi.org/10.1080/00207540110052157.

Güngör, A., Gupta, S.M., 2002. Disassembly line in product recovery. International Journal of Production Research 40(11), 2569–2589. https://doi.org/10.1080/00207540210135622.

Hezer, S., Kara, Y., 2014. A network-based shortest route model for parallel disassembly line b alancing problem. International Journal of Production Research 53(6), 1849–1865. https://doi.org/10.1080/00207543.2014.965348.

Kalayci, C.B., Gupta, S.M., 2013a. Artificial bee colony algorithm for solving sequence-dependent disassembly line balancing problem. Expert Systems with Applications 40(18), 7231–7241. https://doi.org/10.1016/j.eswa.2013.06.067.

Kalayci, C.B., Gupta, S.M., 2013b. A particle swarm optimization algorithm with neighborhood-based mutation for sequence-dependent disassembly line balancing problem. The International Journal of Advanced Manufacturing Technology 69(1-4), 197–209. https://doi.org/10.1007/s00170-013-4990-1.

Kalayci, C.B., Gupta, S.M., 2013c. A tabu search algorithm for balancing a sequence-dependent disassembly line. Production Planning & Control 25(2), 149–160. https://doi.org/10.1080/09537287.2013.782949.

Kalayci, C.B., Hancilar, A., Gungor, A., Gupta, S.M., 2015. Multi-objective fuzzy disassembly line balancing using a hybrid discrete artificial bee colony algorithm. Journal of Manufacturing Systems 37, 672–682. https://doi.org/10.1016/j.jmsy.2014.11.015.

Kalayci, C.B., Polat, O., Gupta, S.M., 2014. A hybrid genetic algorithm for sequence-dependent disassembly line balancing problem. Annals of Operations Research 242(2), 321–354. https://doi.org/10.1007/s10479-014-1641-3.

Li, Y., Kucukkoc, I., Tang, X., 2020. Two-sided assembly line balancing that considers uncertain task time attributes and incompatible task sets. International Journal of Production Research 59(6), 1736–1756. https://doi.org/10.1080/00207543.2020.1724344.

Liang, J., Guo, S., Du, B., Li, Y., Guo, J., Yang, Z., Pang, S., 2021. Minimizing energy consumption in multi-objective two-sided disassembly line balancing problem with complex execution constraints using dual-individual simulated annealing algorithm. Journal of Cleaner Production 284. https://doi.org/10.1016/j.jclepro.2020.125418Get rights and content.

Liang, J., Guo, S., Du, B., Liu, W., Zhang, Y., 2022. Restart genetic flatworm algorithm for two-sided disassembly line balancing problem considering negative impact of destructive disassembly. Journal of Cleaner Production 355. https://doi.org/10.1016/j.jclepro.2022.131708.

Liu, Q., Wei, X., Wang, Q., Song, J., Lv, J., Liu, Y., Tang, O., 2024. An investigation of mixed-model assembly line balancing problem with uncertain assembly time in remanufacturing. Computers & Industrial Engineering 198. https://doi.org/10.1016/j.cie.2024.110676.

McGovern, S.M., Gupta, S.M., 2005. A balancing method and genetic algorithm for disassembly line balancing. European Journal of Operational Research 179(3), 692–708. https://doi.org/10.1016/j.ejor.2005.03.055.

McGovern, S.M., Gupta, S.M., 2006. Ant colony optimization for disassembly sequencing with multiple objectives. The International Journal of Advanced Manufacturing Technology 30(5-6), 481–496. https://doi.org/10.1007/s00170-005-0037-6.

Niknejad, A., Petrovic, D., 2014. Optimisation of integrated reverse logistics networks with different product recovery routes. European Journal of Operational Research 238(1), 143–154. https://doi.org/10.1016/j.ejor.2014.03.034.

Pistolesi, F., Lazzerini, B., Mura, M.D., Dini, G., 2018. EMOGA: A Hybrid Genetic Algorithm With Extremal Optimization Core for Multiobjective Disassembly Line Balancing. IEEE Transactions on Industrial Informatics 14(3), 1089–1098. https://doi.org/10.1109/TII.2017.2778223.

Ren, Y., Zhang, C., Zhao, F., Tian, G., Lin, W., Meng, L., Li, H., 2018. Disassembly line bala

- ncing problem using interdependent weights-based multi-criteria decision making and 2-Optimal algorithm. Journal of Cleaner Production 174, 1475–1486. https://doi.org/10.1016/j.jclepro.2017.10.308.
- Tang, Y., Li, C., 2012. Uncertainty management in remanufacturing: A review. 2012 IEEE International Conference on Automation Science and Engineering (CASE). pp. 52–57. https://doi.org/10.1109/CoASE.2012.6386365.
- Wang, K., Li, X., Gao, L., 2019. Modeling and optimization of multi-objective partial disassembly line balancing problem considering hazard and profit. Journal of Cleaner Production 211, 115–133. https://doi.org/10.1016/j.jclepro.2018.11.114.
- Wang, Y., Hu, F., Wang, Y., Guo, J., Yang, Z., Jiang, F., 2023. Revolutionizing the Afterlife of EV Batteries: A Comprehensive Guide to Echelon Utilization Technologies. ChemElectroChem 11(4). https://doi.org/10.1002/celc.202300666.
- Xiao, Q., Guo, X., Li, D., 2020. Partial disassembly line balancing under uncertainty: robust optimisation models and an improved migrating birds optimisation algorithm. International Journal of Production Research 59(10), 2977–2995. https://doi.org/10.1080/00207543.2020.1744765.
- Xu, W., Cui, J., Liu, B., Liu, J., Yao, B., Zhou, Z., 2021. Human-robot collaborative disassembly line balancing considering the safe strategy in remanufacturing. Journal of Cleaner Production 324. https://doi.org/10.1016/j.jclepro.2021.129158.
- Yang, Y., Yuan, G., Zhuang, Q., Tian, G., 2019. Multi-objective low-carbon disassembly line balancing for agricultural machinery using MDFOA and fuzzy AHP. Journal of Cleaner Production 233, 1465–1474. https://doi.org/10.1016/j.jclepro.2019.06.035.
- Yin, T., Zhang, Z., Wu, T., Zeng, Y., Zhang, Y., Liu, J., 2023. Multimanned partial disassembly line balancing optimization considering end-of-life states of products and skill differences of workers. Journal of Manufacturing Systems 66, 107–126. https://doi.org/10.1016/j.jmsy.2022.12.002.
- Zeng, Y., Zhang, Z., Liang, W., Zhang, Y., 2023. Balancing optimization for disassembly line of mixed homogeneous products with hybrid disassembly mode. Computers & Industrial Engineering 185. https://doi.org/10.1016/j.cie.2023.109646.
- Zeng, Y., Zhang, Z., Yin, T., Zheng, H., 2022. Robotic disassembly line balancing and sequencing problem considering energy-saving and high-profit for waste household appliances. Journal of Cleaner Production 381. https://doi.org/10.1016/j.jclepro.2022.13520.
- Zhou, J., 2023. Research status of new energy vehicles in various countries and the significance of new energy vehicles. Applied and Computational Engineering 12(1), 199–205. https://doi.org/10.54254/2755-2721/12/20230339.